\documentclass[runningheads]{llncs}
\usepackage[T1]{fontenc}
\usepackage{graphicx}
\usepackage{listings}
\usepackage{stmaryrd}
\usepackage{algorithm}
\usepackage{algcompatible}
\usepackage{algpseudocode}

\usepackage{amsmath,amssymb}
\usepackage{subcaption}
\usepackage{stmaryrd}
\usepackage{booktabs}
\usepackage{tabularx}
\usepackage{dirtytalk}
\usepackage{multirow}
\usepackage{adjustbox}
\usepackage{bm}
\usepackage{bbding}

\usepackage{makeidx}  
\usepackage[colorlinks,linkcolor=blue]{hyperref}
\usepackage{url}
\usepackage[misc]{ifsym}

\usepackage[acronym]{glossaries}

\makeatletter
\newcommand*{\inlineequation}[2][]{%
  \begingroup
    \refstepcounter{equation}%
    \ifx\\#1\\%
    \else
      \label{#1}%
    \fi
    \relpenalty=10000 %
    \binoppenalty=10000 %
    \ensuremath{%
      #2%
    }
    ~\@eqnnum
  \endgroup
}
\makeatother

\begin{document}
\title{Aleatoric Uncertainty Medical Image Segmentation Estimation via Flow Matching}
\titlerunning{Uncertainty Flow Matching}

\author{
  Van Phi Nguyen\textsuperscript{*},
  Ngoc Huynh Trinh\textsuperscript{*},
  Duy Minh Lam Nguyen,
  Phu Loc Nguyen,
  Quoc Long Tran\textsuperscript{$\dagger$\ \Letter}
}

\institute{
  Institute for Artificial Intelligence, University of Engineering and Technology, \\
  Vietnam National University, Hanoi, Vietnam \\
  \email{\{phinv, huynhtn, 22022605, 22022547, tqlong\}@vnu.edu.vn} \\
  \textsuperscript{*}Equal contribution \quad\quad \textsuperscript{$\dagger$}Corresponding author
}

\authorrunning{Phi, Huynh et al.}
\maketitle

\begin{abstract}
Quantifying aleatoric uncertainty in medical image segmentation is critical since it is a reflection of the natural variability observed among expert annotators. A conventional approach is to model the segmentation distribution using the generative model, but current methods limit the expression ability of generative models. While current diffusion-based approaches have demonstrated impressive performance in approximating the data distribution, their inherent stochastic sampling process and inability to model exact densities limit their effectiveness in accurately capturing uncertainty. In contrast, our proposed method leverages conditional flow matching, a simulation-free flow-based generative model that learns an exact density, to produce highly accurate segmentation results. By guiding the flow model on the input image and sampling multiple data points, our approach synthesizes segmentation samples whose pixel-wise variance reliably reflects the underlying data distribution. This sampling strategy captures uncertainties in regions with ambiguous boundaries, offering robust quantification that mirrors inter-annotator differences. Experimental results demonstrate that our method not only achieves competitive segmentation accuracy but also generates uncertainty maps that provide deeper insights into the reliability of the segmentation outcomes. The code for this paper is freely available at \url{https://github.com/huynhspm/Data-Uncertainty}

\end{abstract}
\section{Introduction}

Medical image segmentation plays a vital role in diagnosis and treatment planning, making it important to understand how confident a model is in its predictions. Automated segmentation's uncertainty can alert clinicians to ambiguous cases, potentially avoiding misdiagnoses. In clinical practice, even expert annotators often disagree on boundaries or extents, meaning there may be multiple plausible “ground truth” segmentations for a given image~\cite{doi1999computer}. This phenomenon is often called aleatoric (or data) uncertainty. The intrinsic ambiguity in the data that a deterministic model cannot express with a single output and cannot be reduced with more data~\cite{kendall2017uncertainties}. Indeed, studies report low inter-annotator agreement in many medical segmentation tasks, underscoring the need for models that can capture this ambiguity~\cite{joskowicz2019inter}. 

Recent studies have explored a variety of approaches to quantify uncertainty in medical image segmentation~\cite{jungo2019assessing,huang2024review}. Classical methods such as Monte Carlo dropout~\cite{gal2016dropout} and deep ensembles~\cite{lakshminarayanan2017simple} estimate epistemic uncertainty by performing multiple stochastic forward passes through a network. Alternatively, Bayesian neural networks offer a principled approach by treating the network weights as random variables with posterior distributions conditioned on the training data, thus naturally capturing epistemic uncertainty through integration over the weight space~\cite{blundell2015weight}. These methods, however, often focus on epistemic uncertainty, the uncertainty in the model's parameters, and may not fully capture the aleatoric uncertainty. In contrast, generative models, such as conditional variational autoencoders~\cite{sohn2015learning}, address these limitations by learning a joint latent representation of the input images and their segmentation maps. However, current generative models based on diffusion models~\cite{ho2020denoising,song2020score,rahman2023ambiguous} introduce stochasticity at sampling time, leading to unreliable uncertainty mapping. In recent years, conditional flow matching~\cite{lipman2022flow} recently emerged as an alternative method to approximate data distribution by learning continuous, smooth diffeomorphic transforms from a simple distribution into complex data distribution that allow smooth sample generation.

\noindent\textbf{Contribution: } In this work, we introduce a novel method for estimating aleatoric uncertainty in medical image segmentation using a conditional flow matching framework. Unlike diffusion-based models, which approximate the segmentation map distribution and introduce stochasticity during sampling, our approach directly learns an exact, deterministic velocity field conditioned on both the input image and expert annotations. This design allows us to generate segmentation samples that not only align closely with the underlying anatomical context but also accurately capture the inherent variability among multiple expert annotations.

\noindent\textbf{Related work:}
\noindent\textbf{Uncertainty segmentation estimation: } Capturing both epistemic and aleatoric is crucial for modeling the uncertainty in medical segmentation. Bayesian neural networks (BNNs) have been employed to capture epistemic uncertainty by placing priors over model parameters and using variational inference for posterior estimation~\cite{seedat2019towards}. However, BNNs often face challenges with computational complexity and scalability. Monte Carlo dropout~\cite{gal2016dropout} offers a more practical alternative by approximating Bayesian inference through dropout at inference time, enabling uncertainty estimation with minimal computational overhead. A recent advancement, Laplacian Segmentation Networks (LSN)~\cite{zepf2024laplacian}, leverages a Laplacian pyramid to extract multi-scale features and models scale specific priors through independently trained branches, resulting in improved estimation of epistemic uncertainty. In contrast, Aleatoric uncertainty aims to capture the inherent variability of data distribution. Hence, learning the underlying data distribution is often the goal of current approaches. Probabilistic UNet (Prob-UNet)~\cite{kohl2018probabilistic} and its variants PHiSeg~\cite{baumgartner2019phiseg} introduce a latent variable model to generate diverse segmentation hypotheses. Recently, Collectively Intelligent Medical Diffusion (CIMD)~\cite{rahman2023ambiguous} employs a diffusion-based generative process that iteratively corrupts and refines segmentation masks conditioned on the input image, enabling coherent sampling of multimodal hypotheses and robust quantification of segmentation ambiguity. As a continuation of diffusion-based approaches, Conditional Categorical Diffusion Models (CCDM)~\cite{zbinden2023stochastic} extend this idea by introducing a categorical diffusion process that directly operates on discrete label spaces, thereby improving the modeling of multimodal and ambiguous segmentation outputs. Another approach is test-time augmentation~\cite{wu2023upl}, which runs the segmentation model on various transformed versions of the input (rotated, scaled, noised, etc.) and observes the consistency of the outputs. 

\noindent\textbf{Deep Generative Models: } Generative models have significantly advanced medical image segmentation by modeling the data distribution~\cite{wu2024medsegdiff}. Variational Autoencoders (VAEs)~\cite{kingma2013auto} introduced a probabilistic framework that learns latent representations of data, facilitating the generation of diverse segmentation outputs. Despite early success, VAEs soon suffer from blurry outputs due to the inherent trade-off between reconstruction fidelity and latent space regularization. Diffusion models~\cite{song2020score,ho2020denoising} address this by defining a forward process that adds noise to the data and a learned reverse process that denoises it, producing high-quality and diversely generated samples. Conditional flow matching~\cite{lipman2022flow} learning a vector field that maps simple distributions to complex ones without requiring simulation of the probabilistic flow ordinary differential equation (ODE). The deterministic, locally smooth flow also allows generating samples that closely resemble a smooth data manifold and better capture data uncertainty.
    
\section{Method}

\begin{figure}[t!]
    \centering
    \includegraphics[width=\linewidth]{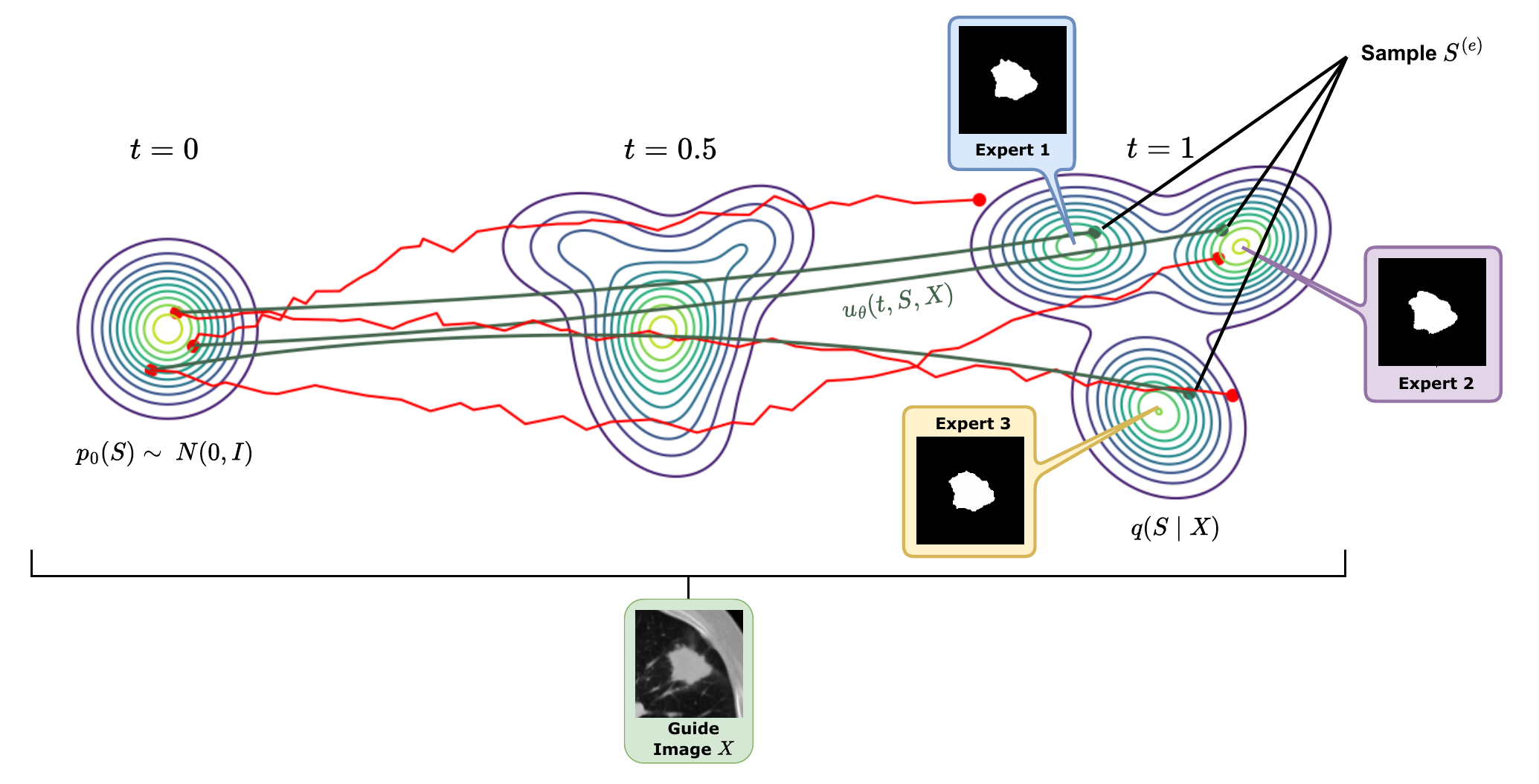}
    \caption{Illustration of our proposed conditional flow matching framework for modeling multi-expert segmentation variability.  At $t=0$, samples $S_0$ are drawn from a simple Gaussian prior $p_0(S)$.  As $t$ progresses, the flow $\psi_t$ (green lines) evolves these samples toward the target distribution $q(S\mid X)$ at $t=1$.  Unlike diffusion methods (red contours) that inject noise and risk blurring fine details, flow matching directly learns a velocity field $u_\theta(t,S,X)$ that preserves local structure.  Multiple expert segmentations (blue, purple, and yellow boxes) are incorporated by conditioning on their annotation.  The guide image $X$ (bottom) provides anatomical context to ensure samples $S^{(e)}$ align closely with the underlying anatomy.}
    \label{fig:narrow_model}
\end{figure}

 \noindent Our goal is to directly model the full segmentation distribution, hence capturing the variability of expert annotations. In standard diffusion models, the forward process adds noise incrementally to the data, and the reverse process is learned via a score function. Although diffusion models are capable of sampling from complex distributions, the injected noise tends to obscure the fine local structure associated with expert uncertainty. In contrast, flow matching directly learns a velocity field conditioned on the image and on local expert annotations, thereby preserving local structure. Moreover, because the flow is designed to interpolate between source and target distribution, the resulting samples more faithfully reflect the true underlying uncertainty-without the potential over-smoothing or blurring that can arise from the stochastic noise in diffusion processes. Formally, we have a medical segmentation dataset
\begin{equation}
\mathcal{D} = \{(X_i, \{S_i^{(1)}, S_i^{(2)}, \ldots, S_i^{(E)}\})\}_{i=1}^{N},
\end{equation}
where \( X_i \) denotes the \(i\)th medical image and \( \{S_i^{(e)}\}_{e=1}^{E} \) are the segmentation maps produced by \(E\) independent experts. Such a multi-expert annotation scheme is crucial for capturing inter-observer variability, which reflects the inherent aleatoric uncertainty present in the segmentation task.

\noindent\textbf{Segmentation via Conditional Flow Matching: } To model the complex and multimodal segmentation distribution \( q(S \mid X) \) conditioned on the image \( X \), we adopt a conditional flow matching framework~\cite{lipman2022flow}. Flow matching is a generative model that constructs a time-dependent probability path \( \{p_t(S \mid X)\}_{t \in [0,1]} \) bridging a simple source distribution \( p_0(S) \) (e.g., a simple, isotropic Gaussian) and the target distribution \( q(S \mid X) \). Specifically, we define a flow \( \psi_t \) that evolves a source sample \( S_0 \sim p_0 \) according to the ODE,
\begin{equation}
\frac{d}{dt}\psi_t(S_0) = u_\theta(t, \psi_t(S_0), X), \quad \psi_t(S_0) = S_t,
\end{equation}
where \( u_\theta(t, S, X) \) is a learnable, time-dependent velocity field conditioned on the image \( X \). To incorporate the variability of expert annotations, we condition the probability path on a particular expert segmentation \( S^{(e)} \). This gives rise to a conditional probability path
\begin{equation}
p_t(S \mid S^{(e)}, X) = \mathcal{N}(S; \, t\, S^{(e)}, \, (1-t)^2 I),
\end{equation}
which satisfies the boundary conditions
\begin{equation}
p_{t=0}(S \mid S^{(e)}, X) = p_0(S), \quad p_{t=1}(S \mid S^{(e)}, X) = \delta(S - S^{(e)}).
\end{equation}
The overall (marginal) segmentation distribution is then obtained by integrating over the expert annotations:
\begin{equation}
p_t(S \mid X) = \int p_t(S \mid S^{(e)}, X) \, q(S^{(e)} \mid X) \, dS^{(e)}.
\end{equation}

\noindent The velocity field \( u_\theta \) is trained via a regression loss that forces it to match the “ground-truth” conditional velocity. In our case, if we sample
\begin{equation}
S_t = tS^{(e)} + (1-t)\epsilon, \quad \epsilon\sim\mathcal{N}(0,I),
\end{equation}
then the corresponding target velocity is given by
\begin{equation}
    u(t, S_t \mid S^{(e)}, X) = \frac{S^{(e)} - S_t}{1-t}.
\end{equation}
Accordingly, we optimize the conditional flow matching (CFM) loss
\begin{equation}
\mathcal{L}_{\mathrm{CFM}}(\theta) = \mathbb{E}_{t, S^{(e)}, X} \left[ \left\| u_\theta(t, S_t, X) - \frac{S^{(e)} - S_t}{1-t} \right\|^2 \right].
\end{equation}

\noindent In order to further align the segmentation with the visual content of the image while preserving the diversity arising from expert annotations, we employ classifier-free guidance~\cite{ho2022classifier}. Specifically, the velocity field is learned in both conditional and unconditional forms, and the final guided velocity is computed as
\begin{equation}
u_\theta^{\text{guided}}(t, S, X) = u_\theta(t, S, X) + w \Big( u_\theta(t, S, X) - u_\theta(t, S, \varnothing) \Big),
\end{equation}
where \( w \) is a hyperparameter that controls the strength of guidance toward the conditional direction. The term \( u_\theta(t, S, \varnothing) \) represents the unconditional velocity, obtained by removing the image conditioning. By training a single network with random conditioning dropout (e.g., with probability 0.5), we enable both conditional and unconditional predictions using shared parameters \( \theta \). This guidance strategy enhances the fidelity of the segmentation maps to the anatomical structures present in \( X \) while preserving the diversity of plausible expert interpretations.

\noindent \textbf{Quantification of Aleatoric Uncertainty: } A central objective of our method is to quantify the aleatoric uncertainty due to the inherent variability among expert annotations. Once the conditional flow matching model is trained, a segmentation map \( S_1 \) is obtained by integrating the guided velocity field:
\begin{equation}
S_1 = \psi_1(S_0; X), \quad \text{with } S_0 \sim p_0(S).
\end{equation}
To capture uncertainty, we generate multiple segmentation samples by drawing different source samples \( \{S_0^{(i)}\}_{i=1}^M \) and propagating each through the learned flow:
\begin{equation}
S_1^{(i)} = \psi_1(S_0^{(i)}; X), \quad i = 1, \ldots, M.
\end{equation}
The variability among the set \( \{S_1^{(i)}\}_{i=1}^M \) is then quantified using the pixel-wise variance, which reflects the variability of annotators.

\section{Experiment Settings}

\noindent\textbf{Data: } We evaluate our method in two datasets, the Lung Image Database Consortium and Image Database Resource Initiative (LIDC-IDRI)~\cite{armato2004lung} and the Multi-Rater Medical Image Segmentation dataset for Nasopharyngeal Carcinoma (MMIS)~\cite{MMIS2024}. The LIDC-IDRI dataset contains 1,018 chest CT scans, where each scan is a 3D volumetric dataset comprising multiple 2D slices. The dataset focuses on lung nodules and includes detailed annotations by up to four experienced radiologists. Following the Prob-UNet~\cite{kohl2018probabilistic} preprocessing pipeline, individual 2D slices are extracted and cropped to \(128\times128\) pixels to focus on regions containing nodules. The final dataset includes 13,508 images for training and 1,588 images for testing. The MMIS dataset comprises MRI scans of 150 subjects across three modalities (i.e., T1, T2, and T1-Contrast), all aligned to a common anatomical space. The gross tumor volumes of nasopharyngeal carcinoma were independently annotated by four senior radiologists with from five to ten years of clinical experience. All slices are preprocessed and cropped to \(128\times128\) pixels, yielding 2,405 training images and 487 testing images.

\noindent\textbf{Implementation: } We adopt the UNet architecture within the flow matching framework~\cite{lipman2022flow} and choose a resolution of 128×128 for both datasets. For the MMIS dataset, three MRI modalities (e.g., T1, T2, FLAIR) are concatenated along the channel dimension to form a multi-modal image prior that is provided as input to the model. During training, a random annotation mask is selected each time a data point is sampled, exposing the model to diverse expert interpretations. To jointly train the network in both conditional and unconditional modes, the image conditioning is randomly dropped with a fixed probability of 0.5. To train the UNet for approximating the velocity field, the Adam optimizer is used with a learning rate of $10^{-4}$, a batch size of 64, and 100,000 iterations. At inference, we used the midpoint method with a fixed step size of 0.01 for stable and efficient ODE integration. A classifier-free guidance scale of 0.3 is applied to modulate the influence of image conditioning and to generate multiple plausible segmentation masks for each input image. All experiments are conducted on an NVIDIA A100 GPU with 80 GB of memory.

\noindent\textbf{Evaluation: } We evaluate our model using the following metrics: (1) Generalized Energy Distance (GED), (2) Average Normalized Cross Correlation (\(S_\textrm{NCC}\)), (3) Maximum Dice Matching (\(D_{\max}\)), and (4) Dice coefficient. The first two metrics measure the similarity between the predicted and ground truth distributions, with GED~\cite{kohl2019hierarchical} quantifying the overall distributional discrepancy, and \(S_\textrm{NCC}\)~\cite{baumgartner2019phiseg} assessing the structural alignment between the set of generated masks and the ground truth masks. In addition, \(D_{\max}\)~\cite{rahman2023ambiguous} evaluates the best case alignment between predictions and annotations by computing the maximum Dice score between each predicted mask and all ground truth masks. The last metric, Dice, is computed as the mean overlap between the two sets of masks. We compare the performance of our model with Prob-UNet~\cite{kohl2018probabilistic} and PHiSeg~\cite{baumgartner2019phiseg}, using the implementation provided in \href{https://github.com/gigantenbein/UNet-Zoo}{UNet-Zoo}, as well as CIMD~\cite{rahman2023ambiguous}. All models are trained on the same preprocessed datasets and generate the same number of 15 sample masks for evaluation.

\section{Results }

\begin{table}[htbp]
\vspace{-1cm}
  \centering
    \caption{Quantitative evaluation of our model on two datasets compared to three baseline methods. The column \(N_\textrm{sample}\) is the number of samples generated to calculate the uncertainty. The best results are highlighted in \textbf{bold}.}
  \label{tab:quantitative}
  \begin{tabular}{l@{\hspace{10pt}}lccccc}
    \toprule
    Dataset & Method & \(N_\textrm{sample}\) & GED $\downarrow$ & \(S_\textrm{NCC} \uparrow\)
 & \(D_{\max}\) $\uparrow$ & Dice $\uparrow$ \\
    \midrule
    \multirow{13}{*}{LIDC-IDRI~\cite{armato2004lung}} 
       & Prob-UNet~\cite{kohl2018probabilistic}  & 5  & 0.353 & 0.750 & 0.787 & 0.505 \\
       &                                      & 10 & 0.300 & 0.817 & 0.859 & 0.520 \\
       &                                      & 15 & \textbf{0.290} & 0.831 & 0.870 & 0.525 \\
       \\
       & PHiSeg~\cite{baumgartner2019phiseg} & 5  & 0.533 & 0.310 & 0.565 & 0.579 \\
       &                                          & 10 & 0.518 & 0.342 & 0.578 & 0.581 \\
       &                                          & 15 & 0.513 & 0.358 & 0.586 & 0.577 \\
       \\
       & CIMD~\cite{rahman2023ambiguous} & 5  & 0.395 & 0.791 & 0.778 & 0.556 \\
       &                                          & 10 & 0.339 & 0.828 & 0.835 & 0.584 \\
       &                                          & 15 & 0.297 & 0.830 & 0.870 & 0.556 \\
       \\
       & Proposed (Ours)                      & 5  & 0.409 & 0.816 & 0.739 & 0.537 \\
       &                                      & 10 & 0.323 & 0.832 & 0.848 & 0.578 \\
       &                                      & 15 & 0.292 & \textbf{0.835} & \textbf{0.876} & \textbf{0.595} \\
    \midrule
     \multirow{13}{*}{MMIS~\cite{MMIS2024}} 
       & Prob-UNet~\cite{kohl2018probabilistic}       & 5  & 0.280 & 0.501 & 0.813 & 0.729 \\
       &                                            & 10 & 0.240 & 0.574 & 0.845 & 0.738 \\
       &                                            & 15 & \textbf{0.227} & 0.607 & 0.856 & 0.740 \\
       \\
       & PHiSeg~\cite{baumgartner2019phiseg}   & 5  & 0.328 & 0.217 & 0.749 & 0.770 \\
       &                                            & 10 & 0.321 & 0.243 & 0.755 & 0.776 \\
       &                                            & 15 & 0.319 & 0.254 & 0.757 & 0.780 \\
       \\
       & CIMD~\cite{rahman2023ambiguous}   & 5  & 0.301 & 0.568 & 0.797 & 0.765 \\
       &                                            & 10 & 0.256 & 0.695 & 0.834 & 0.773 \\
       &                                            & 15 & 0.235 & 0.742 & 0.849 & 0.783 \\
       \\
       & Proposed (Ours)                            & 5  & 0.294 & 0.749 & 0.813 & 0.770 \\
       &                                            & 10 & 0.244 & 0.778 & 0.842 & 0.778 \\
       &                                            & 15 & 0.231 & \textbf{0.789} & \textbf{0.856} & \textbf{0.785} \\
    \bottomrule
  \end{tabular}
\end{table}

Table~\ref{tab:quantitative} presents a quantitative comparison of our proposed method against three baselines: Prob-UNet~\cite{kohl2018probabilistic}, PHiSeg~\cite{baumgartner2019phiseg}, and CIMD~\cite{rahman2023ambiguous} on the LIDC-IDRI and MMIS datasets. Across all evaluation metrics, our approach achieves superior performance, except for GED, where it is only slightly outperformed by Prob-UNet. For the LIDC-IDRI dataset, as the number of generated samples increases from 5 to 15, most methods show improvement across all four metrics. Our method shows a steady improvement, reducing the GED from 0.409 to 0.292 and increasing \(S_{\textrm{NCC}}\) from 0.816 to 0.835. Moreover, \(D_{\max}\) and Dice reach 0.876 and 0.595, respectively, at 15 samples, outperforming all baselines. Similar trends are observed in the MMIS dataset, where our approach attains \(S_{\textrm{NCC}}\) of 0.789, \(D_{\max}\) of 0.856, and Dice of 0.785 with 15 samples, again outperforming the baselines. Despite achieving a GED of 0.231, which is marginally higher than the 0.227 of Prob-UNet, our method still significantly outperforms the other two approaches. These results indicate that our conditional flow matching framework not only produces segmentation maps that closely align with expert annotations, but also effectively captures the inherent aleatoric uncertainty. The ability to generate multiple plausible segmentation hypotheses, with improved fidelity as evidenced by the evaluation metrics, demonstrates that the proposed method can better reflect the variability observed among expert annotators. 

\begin{figure}
\vspace{-0.5cm}
    \includegraphics[width=\linewidth]{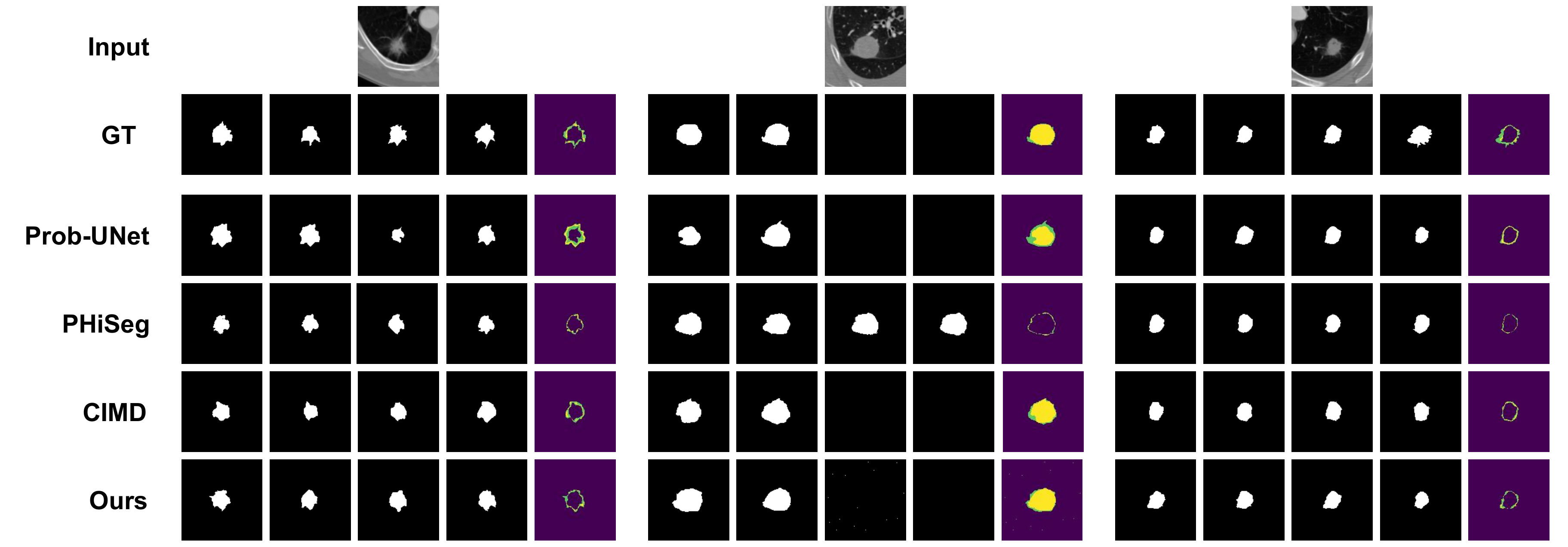}
    \caption{Comparative qualitative analysis of our proposed method against three baseline methods is presented. Three examples from the LIDC-IDRI dataset are shown, with the input images in the first row, followed by rows displaying four segmentation masks and uncertainty maps for the ground truth (GT), the three baseline methods, and our proposed method.}
    \label{fig:qualitative_lidc}
\vspace{-0.5cm}
\end{figure}

\begin{figure}[t!]
    \includegraphics[width=\linewidth]{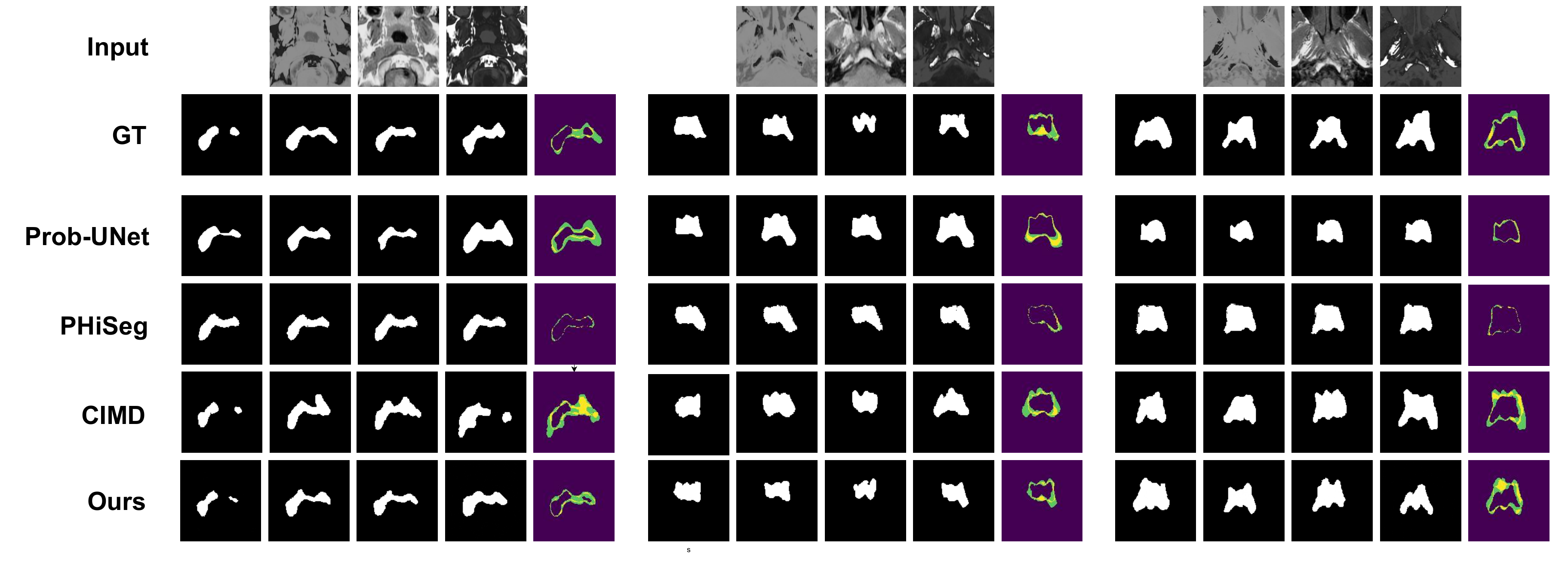}
    \caption{Comparative qualitative analysis of our proposed method against three baseline methods is presented. Three examples from the MMIS dataset are shown, with the input triplet images in the first row, followed by rows displaying four segmentation masks and uncertainty maps for the ground truth (GT), the three baseline methods, and our proposed method.}
    \label{fig:qualitative_mmis}
\vspace{-0.5cm}
\end{figure}

\noindent Figures~\ref{fig:qualitative_lidc} and~\ref{fig:qualitative_mmis} illustrate segmentation outputs generated by our model and baseline methods on challenging cases from the LIDC-IDRI and MMIS datasets. Visually, PHiSeg struggles to learn more than one mask from the ground truth, whereas Prob-UNet and CIMD partially improve by capturing a greater variety of labels. In comparison, our proposed method performs better by effectively generating all ground truth labels, as clearly demonstrated in examples from Figure~\ref{fig:qualitative_mmis}. This highlights our method's capability to effectively learn multimodal distributions under data uncertainty. In addition, the segmentation maps exhibit high fidelity to the anatomical structures present in the images, effectively capturing fine details along the boundaries of the lesions. Notably, in regions where expert annotations disagree, such as fuzzy lesion margins or areas with low contrast, the multiple segmentation samples generated by our method demonstrate a nuanced spread of plausible contours, reflecting its robustness and adaptability.

\noindent\textbf{Limitations: } Despite promising results, our method has some limitations. It estimates only aleatoric uncertainty, neglecting epistemic uncertainty, which arises from the uncertainty of model parameters. This could limit the method's reliability in cases of limited training data or out-of-distribution samples. Incorporating epistemic uncertainty could improve robustness in such scenarios. Additionally, our sampling strategy, which involves randomly drawing samples and solving the ODE integration via the midpoint method, is sensitive to noise levels and computationally demanding, especially for high-resolution images. Exploring more efficient sampling and integration strategies could address these challenges and enhance the method's scalability.

\section{Conclusion} We introduced a novel method for estimating aleatoric uncertainty in medical image segmentation using a conditional flow matching framework. Unlike stochastic diffusion-based approaches, our method directly learns a velocity field, preserving fine anatomical details while capturing variability in multi-expert annotations. Experiments on the LIDC-IDRI and MMIS datasets demonstrated that our approach outperforms previous methods. We also show that the segmentation maps closely capture the variance observed in the ground truth, providing doctors with reliable results for more in-depth analysis. Future work will focus on incorporating epistemic uncertainty, developing more advanced sampling techniques beyond random noise to enhance robustness, and exploring potential clinical applications.

\subsubsection*{Acknowledgment} 
The computing resources used in this research were sponsored by Intelligent Integration Co., Ltd. (INT2), Vietnam.

\subsubsection*{Disclosure of Interest}
The authors declare that they have no competing interests relevant to the content of this article.

\newpage
\bibliographystyle{splncs04}
\bibliography{Paper-1171.bib}

\end{document}